# Expert-level vision-language foundation model for real-world radiology and comprehensive evaluation

Xiaohong Liu[1,2], Guoxing Yang[2], Yulin Luo[3], Jiaji Mao[1], Xiang Zhang[1], Haibo Wang[4], Zhiyang He[5], Ming Gao[1], Shanghang Zhang[3], Jun Shen[1], Guangyu Wang[2]

*[1]Department of Radiology, Sun Yat-Sen Memorial Hospital & Guangdong Provincial Key Laboratory of Malignant Tumor Epigenetics and Gene Regulation, Medical Research Center, Sun Yat-Sen Memorial Hospital, Sun Yat-Sen University, Guangzhou, China*
*[2]State Key Laboratory of Networking and Switching Technology, Beijing University of Posts and Telecommunications, Beijing, China*
*[3]National Key Laboratory for Multimedia Information Processing, School of Computer Science, Peking University, Beijing, China*
*[4]Research Centre of Big Data and Artificial Intelligence in Medicine, The First Affiliated Hospital of Sun Yat-sen University*
*[5]Xunfei Healthcare Technology Co., Ltd.*

[*]These authors contributed equally to this work.

[#] Corresponding authors:
Guangyu Wang, PhD
Email address: guangyu.wang24@gmail.com

## Abstract

Radiology is a vital and complex component of modern clinical workflow and covers many tasks. Recently, vision-language (VL) foundation models in medicine have shown potential in processing multimodal information, offering a unified solution for various radiology tasks. However, existing studies either pre-trained VL models on natural data or did not fully integrate vision-language architecture and pretraining, often neglecting the unique multimodal complexity in radiology images and their textual contexts. Additionally, their practical applicability in real-world scenarios remains underexplored. Here, we present RadFound, a large and open-source vision-language foundation model tailored for radiology, that is trained on the most extensive dataset of over 8.1 million images and 250,000 image-text pairs, covering 19 major organ systems and 10 imaging modalities. To establish expert-level multimodal perception and generation capabilities, RadFound introduces an enhanced vision encoder to capture intra-image local features and inter-image contextual information, and a unified cross-modal learning design tailored to radiology. To fully assess the models' capability, we construct a benchmark, RadVLBench, including radiology interpretation tasks like medical vision-language question-answering, as well as text generation tasks ranging from captioning to report generation. We also propose a human evaluation framework. When evaluated on the real-world benchmark involving three representative modalities, 2D images (chest X-rays), multi-view images (mammograms), and 3D images (thyroid CT scans), RadFound significantly outperforms other VL foundation models on both quantitative metrics and human evaluation. In summary, the development of RadFound represents an advancement in radiology generalists, demonstrating broad applicability potential for integration into clinical workflows.

## Introduction

Radiology is a vital component of modern clinical workflow and plays a crucial role in disease detection, differential diagnosis, and monitoring treatment response across clinical practices. The clinical workflow of radiologists involves a large range of tasks, ranging from the interpretation of multi-modality images such as X-ray, CT, and MRI, to the formulation of text description with possible diagnoses[1,2]. Recently, artificial intelligence (AI) and vision-language models (VLM)[3] have advanced rapidly in the field of radiology, demonstrating their potential to mimic a physician's ability to process multimodal information and improve diagnostic accuracy[4]. However, these early methods were primarily designed for single tasks utilizing a supervised learning approach trained on specific annotated datasets, lacking interactive capabilities, thus limiting their generalizability in real-world clinical workflow.

To address these issues, attention has recently been devoted to vision-language (VL) foundation models in medicine[5], which are typically pretrained on extensive image-text pairs, and promise to provide a unified solution to various downstream tasks. For example, PMC-VQA[6] and LlaVA-Med[7] have demonstrated their effectiveness in visual question answering tasks, while models like XrayGPT[8], R2GenGPT[9], and CXR-Flamingo[10] have shown potential in generating radiological reports for chest X-rays The potential utility of VL foundation models in radiology is extensive, as they can generate tailored text outputs from medical images, and enable natural language interactions with specific users, which might ultimately improve access to diagnostic and prognostic expertise.

However, current VL foundation models are in the preliminary stages of achieving an expert-level solution for real-world radiology. Existing approaches are often pre-trained on general natural images[11,12] or natively trained using medical images[13], and perform cross-modal learning to rely on large language models' strong generation capability, which lacks specifically designed vision encoders for multimodal medical images in this aspect. Further, to bridge the image-text modality gap in radiology, recent efforts[13,14] simply apply state-of-the-art approaches, such as model architectures design (e.g., BLIP, InstructBLIP), or pretraining strategies like Flamingo[15] that adopt interleaved image-text data. However, these approaches often yield suboptimal performance due to oversights in addressing the unique multimodal complexity in radiology which requires unified architectural and pretraining innovations for diverse radiological images.

To address the challenges, we propose RadFound, a medical vision-language foundation model that achieves expert-level multimodal understanding and generation to adapt to diverse tasks encountered in clinical contexts (**Fig. 1a**). RadFound is built on the BLIP-2 architecture[16] with a medical large language model (LLM), an innovative pretrained vision encoder RadVision, and a pretrained vision-language aligner RadFormer. It is conducted on a curated and global peer-reviewed dataset RadVLCorpus which comprises 8.1 million images and over 250,000 image-text pairs, covering 19 organs or systems and 10 modalities to ensure a diverse and representative dataset (**Fig. 1b and 1c**). Each image is accompanied by marked tags, such as information on the corresponding view, modality, organ types. These tags can be utilized as context information, which is key to endowing them with interleaved image-text multimodal representation learning capabilities.

To establish multimodal perception and reasoning capabilities, it is key to facilitate enhanced vision encoders and cross-modal alignment strategies specifically tailored to radiology. Visual perception is the initial step in a radiologist's work, aiming to identify discriminative patterns and understand the holistic context of medical images. First, inspired by the way radiologists analyze medical images, RadVision introduces a novel vision encoder pretraining paradigm, called contextualized contrastive masked image modeling (CC-MIM). This approach combines masked image modeling which captures intra-image local sensitive features, with contextual-based contrastive learning for inter-image contextual information. For vision-language alignment, RadFormer introduces a unified design that incorporates an interleaved image-text data augmentation (ITA) strategy for VL pretraining approach and structural design of multi-image-based instruction-aware module, tailored for multimodal representation for radiology (Fig. **1b**). To fully evaluate the VL foundations in clinical use, we conducted a stepwise evaluation framework based on our proposed RadVLBench , including: 1) radiology interpretation tasks (medical vision-language question-answering) ranging from single image-based to more challenging context-rich tasks; 2) text generation tasks from captioning to more complex report generation; 3) real-world radiology tasks covering various modalities and diseases derived from clinical practice; and 4) a human expert assessment framework, which set a new benchmark in the field. Through comprehensive validation, our method demonstrates superior performance compared with existing foundation models and expert radiologists (**Fig. 1d and 1e**).

## Results

## System overview and dataset characteristics

In this study, we present RadFound, a general-purpose vision-language foundation model tailored for radiology, which is developed based on the modular architecture of BLIP-2[16], a state-of-the-art vision-language framework, comprising a medical large language model (LLM), a vision encoder, and a vision-language alignment module to bridge various components. Here, we pretrain it on the RadVLCorpus dataset. To achieve an expert-level vision-language foundation model in radiology, our approach makes several contributions.

**Vision encoder:** We have developed a novel approach, termed CC-MIM, aimed at cultivating an expert-level vision learner. The CC-MIM is composed of two primary components: Masked Image Modeling (MIM) and Inter-Image Contextualized Contrastive Learning (ICC). In the MIM component, the model reconstructs images from randomly masked segments of image patches, enabling it to learn rich and comprehensive representations. Simultaneously, the ICC component employs an auxiliary Transformer to learn contextualized view representations by distinguishing similarities and differences in various contexts, enhancing the model's ability to analyze interrelations among medical images. By combining MIM and ICC, the vision encoder acquires radiologist-level capabilities, which learns both locally sensitive features and a holistic, contextualized understanding. Additionally, we incorporate an adaptive tokenizing module into the vision encoder for processing visual representations from input medical images of varying resolutions and modalities such as 3D formats.

**Vision-language cross-modality learning in radiology:** We propose a uniform design that significantly enhances cross-modality learning in the field of radiology. This design incorporates two primary elements: image-text data augmentation (ITA) pretraining and a multi-image-based instruction-aware module. Our ITA method organizes image-text pairs by semantic similarity, such as modality or disease context, and provides extensive training data with interleaved image-text formats, thus enhancing the VLM's capabilities in clinical contexts. The multi-image-based instruction-aware module built upon the Q-Former architecture, which composes of cross-attention blocks to extract task-related visual representations from an array of radiological images guided by textual instructions. These collectively enhance our method's ability to better learn cross-modal representations, thereby improving its applicability and performance in the context of radiology.

**Stepwise evaluation of VLMs for real-world radiology:** We have established a specialized benchmark, RadVLBench, to comprehensively evaluate the vision-language understanding and generation capabilities of RadFound. This benchmark encompasses thirteen datasets covering visual question-answering tasks (RadVLBench-VQA), text generation tasks (RadVLBench-Cap), and an external validation benchmark from real-world clinical scenarios (RadVLBench-RW).

For RadVLBench-VQA, we incorporated six diverse datasets, including RAD-VQA[17], Med-VQA[18], SLAKE[19], MIMIC-CXR-VQA[20], MedpixQA, and Medical-Diff-VQA[21]. These datasets vary in format and complexity, ranging from single image-based to multiple image-based question answering tasks. For RadVLBench-Cap, we assembled four datasets: ROCO[22], MedpixCap, IU-Xray[23], and MIMIC-CXR[24] (**Fig. 1d**). The ROCO and MedpixCap datasets primarily focus on the generation of concise image captions, while the IU-Xray and MIMIC-CXR datasets involves the generation of detailed radiology reports, which require a deeper level of analysis and thus pose more challenges than simple image captions. The RadVLBench-RW consists of three datasets: R2Chest (2D chest X-ray images), R2Mammo (multi-view images for mammograms), and R2Thyroid (3D thyroid CT scans).

## Performance on medical visual question answering

Medical visual question answering (VQA) tasks are commonly used to assess the understanding capabilities of vision-language models. Within these tasks, models are tasked with generating responses to questions derived from specific medical images. The accuracy of these responses is then meticulously compared against a set of verified ground truth answers. We conducted an extensive evaluation of the VQA capabilities of RadFound by the RadVLBench-VQA ranging from basic perception tasks (e.g., discriminating normal versus abnormal findings) to more complex analysis (e.g., comparison with prior images) (**Fig. 2a**). We benchmarked RadFound against Med-Flamingo[14], a large-scale medical VLM pre-trained on biomedical literature, and other state-of-the-art (SOTA) approaches in the field.

In the RadVLBench-VQA, RadFound consistently outperformed Med-Flamingo across all tasks (**Fig. 2c**). Specifically, on the single-image-based VQA, RadFound achieved accuracies of 86.8%, 74.6%, 86.0%, and 70.6% for RAD-VQA, Med-VQA, SLAKE, and MIMIC-CXR-VQA, marking improvements over Med-Flamingo by 15.1%, 10.8%, 1.2%, and 10.6%, respectively. An analysis of RadFound's performance across various imaging modalities within the SLAKE dataset demonstrated its robustness, with notable performance gains over Med-Flamingo in X-ray, MRI, and CT modalities by 1.7%, 2.6%, and 4.0%, respectively (**Supplementary Figure 1a**). We also observed that RadFound excelled across all categories, including 'modality' and 'plane', achieving accuracies exceeding 90%, which highlights its versatility in handling diverse radiological tasks (**Supplementary Figure 1b**). The Medical-Diff-VQA and MedpixQA datasets evaluated the model's proficiency in advanced radiological tasks, such as comparative analysis and detailed finding characterizations. We utilized ROUGE-L[25] as the metric, which takes into account textual similarity. RadFound achieved a ROUGE-L score of 0.627 on the Medical-Diff-VQA dataset, significantly outperforming Med-Flamingo's score of 0.392. (**Fig. 2c**). As the complexity increased with more images per query, RadFound consistently demonstrated superior performance, whereas Med-Flamingo struggled, showing a marked decline (**Fig. 2e**). On the exam-style MedpixQA dataset, RadFound achieved an accuracy of 70.7%, significantly outperforming Med-Flamingo, which had an accuracy of 32.7%, nearly 40% lower than RadFound (**Fig. 2c**). RadFound outperforms Med-flamingo across all seven organ systems, including the spine, skull, face, heart, gastrointestinal tract, genitourinary system, vascular system, lung, musculoskeletal system, and breast (**Fig. 2d**). Further analysis on question complexity, measured by word length, demonstrated that RadFound maintained high accuracy even with longer questions, showcasing its advanced capabilities in handling complex clinical inquiries (**Supplementary Figure 1c)**.

## Performance on medical image captioning and report generation

Medical text generation tasks represent another challenging for VLMs, which requires the generation of text descriptions based on visual inputs and accompanying textual contexts. Therefore, it is useful to assess the model's capability to recognize visual contents and generate coherent, detailed descriptions for clinical scenarios. We evaluated the text generative capability of RadFound on RadVLBench-Cap (**Fig. 3a**). The ROCO and MedpixCap datasets, are designed for image-captioning tasks (**Fig. 3b**), while the IU-Xray and MIMIC-CXR datasets are used for radiological report generation that include 'findings' and 'impression' sections (**Fig. 3b**). The quality of the generated text was assessed by comparing it to reference texts using the BLEU-4[11], ROUGE-L[25], and METEOR[20] metrics.

In the RadVLBench-Cap, the overall results illustrate RadFound exhibits a significant performance improvement compared to Med-Flamingo (**Fig. 3c and Fig.**

**3d**). Specifically, on the ROCO dataset, RadFound achieved ROUGE-L and METEOR scores of 0.200 and 0.100, respectively, while Med-Flamingo scored only 0.141 and 0.074 (**Fig. 3c)**. Similarly, RadFound attained ROUGE-L and METEOR scores of 0.373 and 0.213 on the MedpixCap dataset, surpassing Med-Flamingo's scores of 0.115 and 0.069. An examination of RadFound's performance across various organ systems within the MedpixCap dataset revealed that RadFound outperformed Med-Flamingo across all organs, showing a significant improvement (**Supplementary Figure 2a**).

Radiological report benchmarks typically feature longer text lengths and contain more professional and detailed descriptions. Here, RadFound was evaluated for its synthesis skills in generating radiological text. On the IU-Xray dataset, RadFound achieved ROUGE-L and BLEU-4 scores of 0.357 and 0.186 respectively, while the baseline model achieved 0.294 and 0.125. On the MIMIC-CXR dataset, RadFound achieved ROUGE-L of 0.306 and BLEU-4 of 0.114, outperforming the baseline model, which obtained noticeably lower scores of 0.219 and 0.050, respectively (**Fig. 3d)**. The excellent performance indicates that RadFound has developed proficient text generation capabilities, including the generation of longer and more complex radiological reports. Additionally, on the MIMIC-CXR dataset, we evaluated the clinical efficacy[26], which measures the accuracy of descriptions for clinical abnormalities across diseases, such as pneumonia and edema (**Supplementary Figure 2b**). The accurate performance of RadFound reflects its capability to generate clinically effective radiological reports.

To further evaluate the model's performance in multimodal tasks and real clinical scenarios, we conducted an analysis using the MIMIC-Multi-Visit[21] dataset, comprising cases from patients with multiple visits. Four experimental settings were used to assess performance under different multimodal conditions: without previous reports and images, with previous reports, with previous images, and with both previous reports and images. As shown in **Fig. 3f and 3g**, the results indicate that RadFound consistently outperforms Med-Flamingo across all settings. Specifically, in the setting with both previous reports and images, RadFound achieved a ROUGE-L score of 0.282 and a BLEU-4 score of 0.096, while Med-Flamingo reached 0.262 and 0.057, respectively. The enhancements highlight RadFound’s robust performance in multimodal analysis, which can potentially be attributed to the module of multimodal vision-language alignment, thus producing informative text descriptions.

**Performance in real-world clinical settings for multimodal radiology tasks**

To further validate the clinical utility of VLMs in real-world clinical settings, we constructed RadVLBench-RW that comprises three radiology image-report datasets across different modalities: 2D chest radiography (R2Chest), multi-view mammography (R2Mammo), and 3D thyroid CT scans (R2Thyroid), with "R2" denoting “radiology report”.

We evaluated the performance of RadFound on the R2Chest, R2Mammo, and R2Thyroid datasets using the ROUGE-L and BLUE-4 (**Fig. 4b**). Our model significantly outperforms Med-Flamingo on three datasets. Specifically, in the R2Chest dataset, RadFound achieves performance comparable to Med-Flamingo despite using only 13% of the training data. Notably, RadFound significantly outperforms Med-Flamingo in the R2Mammo dataset; It surpasses Med-Flamingo's results with less than 10% training data. For the R2Thyroid dataset with 3D images, using only 14% of the training data, RadFound achieves comparable ROUGE-L performance to Med-Flamingo which trained on the entire training set. Similar results were also observed in the BLEU-4 metrics. This demonstrates RadFound’s effectiveness across various radiology imaging scenarios.

***Performance evaluation based on human evaluation framework***

Previous efforts in evaluation models for report generation have primarily relied on natural language generation metrics such as BLEU or ROUGE scores. While useful for assessing linguistic attributes, these metrics are inadequate for measuring the clinical quality of the reports in real-world settings, where clinical relevance are more critical than the textual similarity to reference reports. Here, we proposed a systematic human evaluation framework with multi-dimensional metrics: ‘Readability’, ‘Medical reasoning and consensus’, ‘Missed findings and abnormality’, ‘Misdiagnosis’, ‘Bias’, and ‘Clinical acceptability’, offering a criteria commonly employed to assess the radiologists' expertise in radiological interpretation. We compared the performance between the AI model and human experts on the RadVLBench-RW benchmark. In averaging six evaluation metrics, the results demonstrated that RadFound significantly outperformed the baseline, achieving performance comparable to senior-level radiologists (**Fig. 5)**. In contrast, Med-flamingo consistently demonstrated inferior performance. The results indicate that RadFound’s performance on the chest and mammography datasets slightly exceeded that of junior radiologists and reached comparable performance with senior radiologists. For the thyroid dataset, RadFound achieved comparable performance to that of junior radiologists.

## Discussion

In this study, we introduced a medical vision-language foundation model, RadFound, and evaluated its generalizability in adapting to diverse downstream tasks. RadFound was pre-trained on an extensive dataset of radiology images and texts through a vision pretraining approach and a vision-language alignment approach. The vision pretraining approach not only adapted the vision encoder to medical domain but also enhanced the vision encoder's ability to capture local sensitive features and relationships among medical images. The vision-language alignment pretraining approach enabled the model to bridge between modalities, facilitating the model's capabilities of multimodal representation and descriptive generation. Within RadFound, an adaptive tokenizing module was introduced that could adapt to diverse medical multimodal inputs, and a multi-image instruction-aware module was employed that further enhanced RadFound's understanding and reasoning with multimodal data. When evaluated on the RadVLBench-VQA, RadVLBench-Cap, and RadVLBench-RW, which encompasses an array of VQA tasks and text generation tasks, RadFound demonstrated superior performance across different modalities and conditions. Furthermore, comparative studies between RadFound and radiologists highlighted the model's potential to reach or exceed expert capabilities in generating radiology reports. With a human evaluation framework, it demonstrated RadFound's superior performance in various aspects of practical radiology applications.

Our study has several limitations. Firstly, for VLM radiology tasks, there is a need for further exploration of evaluation metrics. Previous automatic evaluation metrics cannot reflect the clinical utility of the generated reports. While we conducted an expert evaluation that was more accurate and authoritative, it was also costlier and not feasible on a large scale. Therefore, developing a reliable and efficient automatic evaluation metric that aligns with clinicians' preferences is key to constructing medical foundation models. Furthermore, the integration of VLMs into clinical practice requires more extensive investigation. Although our comparative study evaluated the model's performance against expert assessments and analyzed its potential clinical effectiveness, the actual performance of the model within clinical workflows and in collaboration with radiologists may differ from these controlled evaluations. A thorough exploration is necessary to understand its practical impact.

## Methods

### Pretraining dataset

In this study, we curated a large and diverse collection of radiological images and text for the pretraining of RadFound. The Radiology Image-text Corpus dataset (RadVLCorpus) was curated from a combination of several publicly available datasets and radiology reports collected from multi hospitals. All data underwent the following cleaning steps during collection: (1) exclusion of image-text pairs that may contain sensitive information; (2) removal of duplicate image-text pairs; (3) removal of special tags and characters in the text, with standardization of text content formats. Adhering to our data inclusion and cleaning criteria, we ultimately obtained over 8.1 million radiological images from 250,000 imaging series with descriptions, across 19 organs or systems and 10 modalities. It's worth noting that the dataset encompasses 2D images and 3D images within each case. Pre-training on this comprehensive dataset can enhance the model's generalizability for dealing with downstream tasks involving various modalities. Furthermore, the dataset contains tags (such as modality and organ type) marked by a BiomedCLIP for each medical image, including information on the corresponding modalities and organs. Specifically, we utilize BiomedCLIP for zero-shot classification, enabling us to classify new and unseen medical images in RadVLCorpus without the need for extensive labeled data. This facilitates accurate retrieval of similar cases in subsequent steps to help model training. We utilized radiological images and radiological image-text pairs for pretraining.

### Benchmark for the evaluation

Here, we present RadVLBench, a benchmark combining six medical vision-language question-answering (denoted as RadVLBench-VQA), four text generation datasets (denoted as RadVLBench-Cap), and three report generation datasets from real-world data (denoted as RadVLBench-RW), across various settings such as along modalities, number of images, and text length, assessing comprehensive capability.

To provide a comprehensive evaluation of VQA performance in the medical domain, we validated our approach using six datasets, the RAD-VQA[17], Med-VQA[18], SLAKE[19], MIMIC-CXR-VQA[20], Medical-Diff-VQA[21], MedpixQA. For text generation tasks, we conducted experiments using four datasets: the Radiology Objects in COntext (ROCO) dataset[22], Medpix dataset belong to the category of image caption datasets, the IU-Xray dataset[23] and the MIMIC-CXR dataset[24]. All datasets of text generation were followed by the official data division for training and testing. RadVLBench-RW is a comprehensive benchmark designed for the evaluation of radiology report generation models, to assess the transferability and clinical utility of these models in diverse real-world settings. It encompasses three distinct datasets named R2Chest, R2Mammo, and R2Thyroid, each focusing on different radiological modalities including chest X-rays, mammograms, and thyroid CT scans, respectively, acquired from hospitals.

**Baselines**

We compared our method with the general vision language large model, OpenFlamingo-9B[27] and the medical vision language large model, Med-Flamingo[14]. The original Open-Flamingo and Med-Flamingo accomplish downstream tasks through in-context learning, without being finetuned on these tasks. For a fair comparison, we fine-tuned both models on the same training data and evaluated them with the same evaluation settings.

**RadFound**

RadFound is a vision-language foundation model, designed for multimodal understanding and reasoning, which adapts to diverse data encountered in medical contexts and delivers expert-level interpretation in radiology. The RadFound comprises three main components: a language model, a vision encoder, and a vision-language alignment module. RadFound employs an adaptive tokenizing module for processing all forms of multimodal medical image inputs uniformly and Q-Former integrated a multi-image based instruction-aware module for vision-language alignment. The pretraining of RadFound includes two processes: vision pretraining and vision-language alignment pretraining. For the vision encoder, we implemented a contextualized contrastive masked image modeling pretraining framework (CC-MIM). This approach is designed to adapt the vision encoder to the medical domain and to significantly enhance its ability to interpret and analyze representations of complex contextual relationships among medical images, such as multiple views and modalities. In the vision-language alignment pretraining process, we utilized conditional language modeling and interleaved image-text data augmentation (ITA) to improve the model's capabilities in multimodal encoding and reasoning by pretraining on organized data. We organized related pretraining data into an image-text interleaved format by leveraging the tags (such as modality and organ type).

We proposed an adaptive tokenizing module in the vision encoder for unified processing. It standardizes the input images into uniformed sub-images matching predefined sizes. Specifically, for the processing of multi-view and multimodal data, we input images sequentially. The view or modality tags are appended in the text before the image placeholder tokens, providing contextual cues. To adopt textual instruction and multiple images as input, we introduced a multi-image technique into the instruction-aware Q-Former. It adds a self-attention module for query tokens to integrate information from instructions, which encourages the extraction of task-relevant image features. The CC-MIM framework incorporates masked image modeling (MIM) and inter-image contextualized contrastive learning (ICC). Masked Image Modeling (MIM) primarily focuses on reconstructing local content within an individual image, capturing fine-grained details crucial in medical diagnostics. And we integrate Contrastive Learning (CL) into our framework to encourage the model to learn discriminative features by contrasting different images, leading to strong capability in inter-image relationship modeling. For the visual-language pretraining process, we used a conditional language modeling objective and interleaved image-text data augmentation (ITA) on medical image-text pairs. Based on medical image-text pairs, interleaved image-text data is generated by data augmentation. We employ an interleaved image-text data augmentation strategy with marked tags.

**Study design for performance comparison of AI versus radiologists**

We compared the diagnostic accuracy between the AI system and human radiologists. We selected cases of thyroid, breast, and chest x-ray, to compare the performance of our model with that of human radiologists in composing diagnostic reports. Six radiologists were invited and assigned to two groups: group 1, junior radiologists with over 5 years of clinical experience, and group 2, senior radiologists with over 10 years of clinical experience. The models generate reports for all cases. Then these reports are evaluated by an expert panel consisting of three radiologists with 15-20 years of clinical experience, including Readability, Missed findings and abnormality, Misdiagnosis, Medical reasoning and consensus, Bias, and Clinical acceptability. Each report is assessed by radiologists from the expert panel and scored on a scale of 1-5.

**Statistical analysis**

In this study, for the visual question answering (VQA) tasks, we employed accuracy and F1 score as evaluation metrics, following existing work[28]. For the Medical-Diff-VQA dataset, we extended our evaluation to include text-based metrics such as BLEU-4, ROUGE-L, and METEOR, following the original Medical-Diff-VQA paper[21]. For caption generation and CXR report generation tasks, we utilized BLEU-4, ROUGE-L, and METEOR for evaluation. Additionally, we introduced clinical efficiency assessment for CXR report generation using CheXbert[29]. The 95% confidence intervals were derived using bootstrap resampling with 1,000 samples.

## Figure Legend

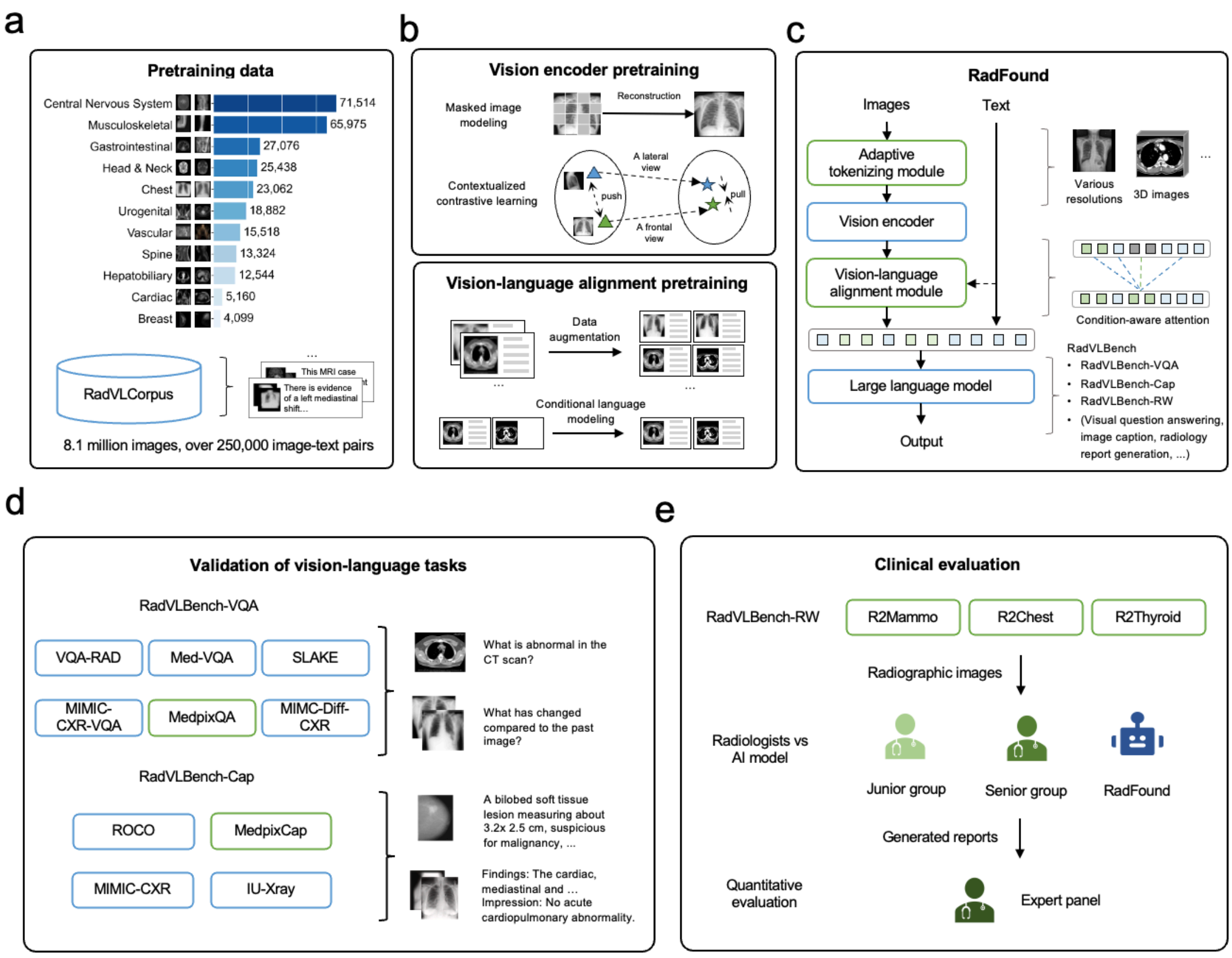

**Figure 1. Schematic illustration of the development and evaluation of RadFound.**

**a,** A radiology image repository, RadVLCorpus, was used as pretraining data, containing radiology images across various modalities and organ categories with text-image pairs. **b,** Vision pretraining and vision-language alignment process. The vision pretraining involved two objectives: Masked Image Modeling (MIM) and Inter-image Contextualized Contrastive Learning (ICC). **c**, The architecture of RadFound. The model consisted of a vision encoder, a vision-language alignment module, and a large language model (LLM). **d,** Benchmark of vision-language tasks. An array of established (blue) and newly introduced (green) datasets for evaluating the performance of vision-language models on radiology-related tasks, including visual question answering (RadVLBench-VQA) and text generation (RadVLBench-Cap). **e,** Clinical evaluation. The study assessed the performance of vision-language models through clinical evaluation, including a comparison of the reports generated by the models with those produced by radiologists at different levels of expertise and a quantitative evaluation across three radiology report datasets (RadVLBench-RW).

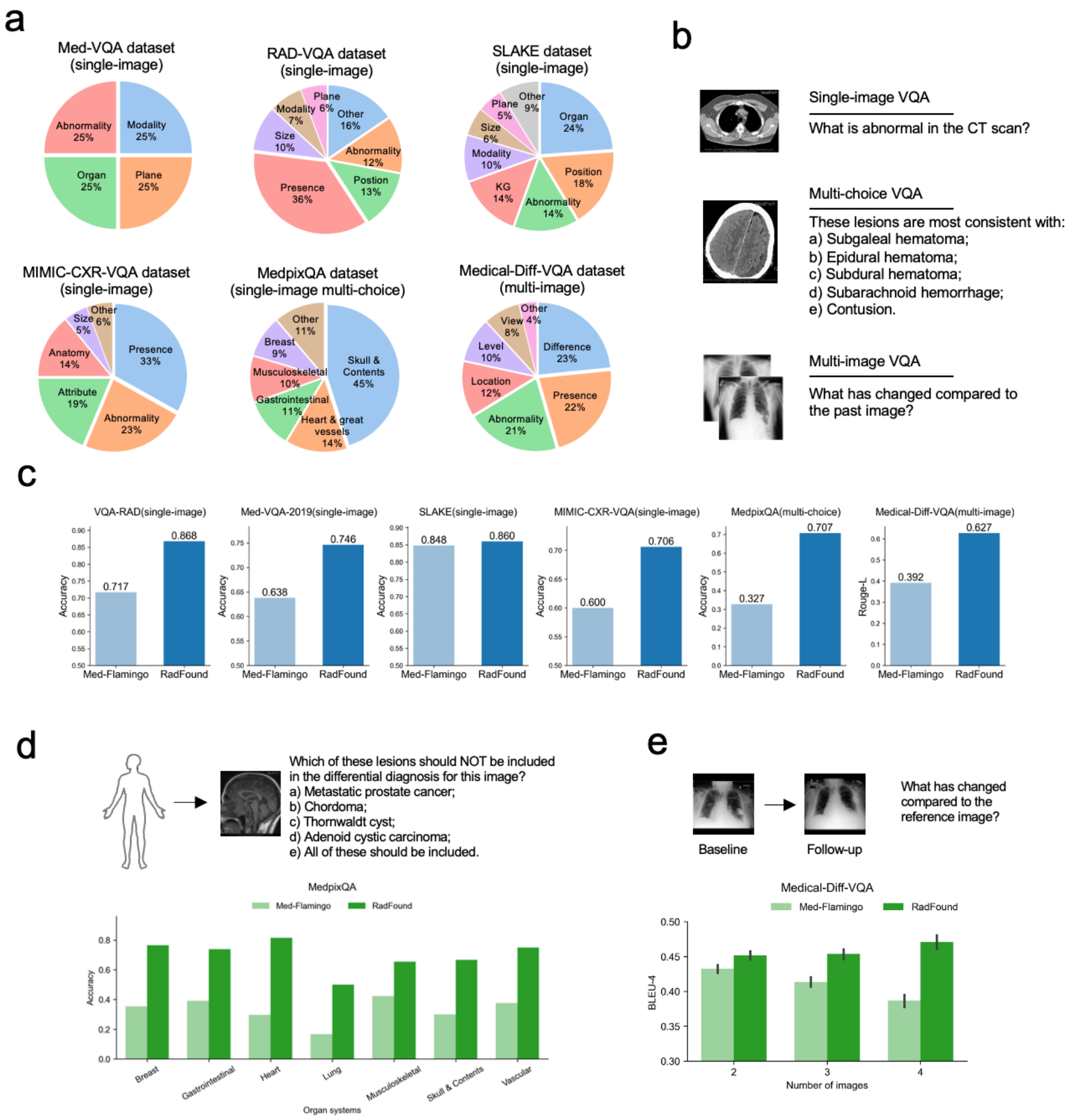

**Figure 2. Performance of model for medical visual question answering.**

**a,** Six medical visual question-answering (VQA) datasets were used for evaluation: RAD-VQA, Med-VQA, SLAKE, EHRXQA, MedpixQA, and Medical-Diff-VQA. **b,** Different types of medical VQA cases, including single-image VQA, single-image multi-choice VQA, and Multi-image VQA. **c,** Performance comparison between Med-Flamingo and RadFound across the six medical VQA datasets. Each bar represents accuracy or ROUGE-L. Dark blue bar and light blue bar represent the performance of RadFound and Med-Flamingo, respectively. **d,** Performance analysis of the accuracy across organs between RadFound and Med-Flamingo on the MedpixQA dataset. Dark green bar and light green bar represent the performance of RadFound and Med-Flamingo, respectively. **e,** Performance analysis of the BLUE-4 score versus the number of images between RadFound and Med-Flamingo on the Medical-Diff-VQA dataset. Dark green bar and light green bar represent the performance of RadFound and Med-Flamingo, respectively.

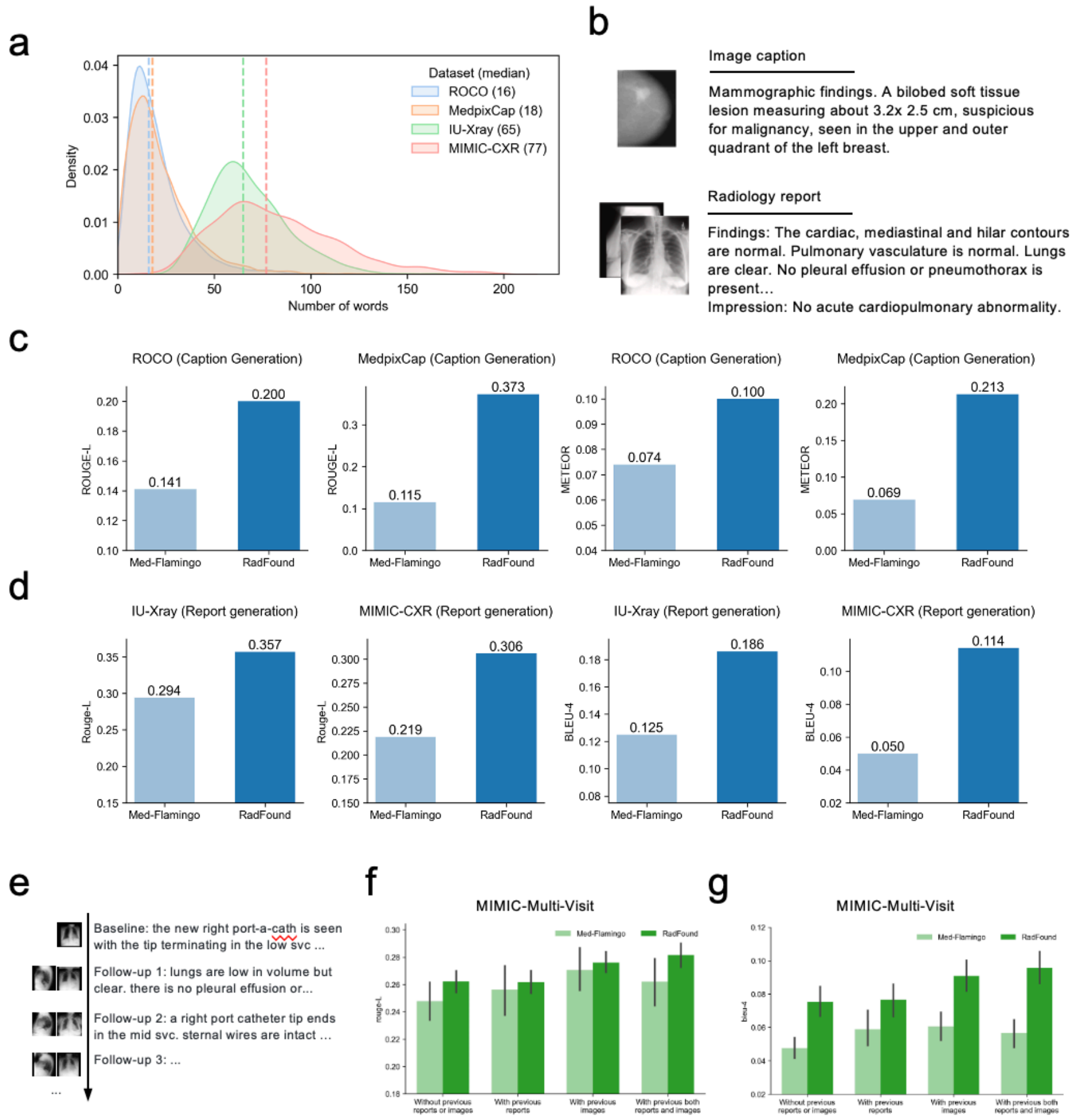

**Figure 3. Performance of models for medical text generation from captioning to report generation.**

**a,** The distribution of textual length across four datasets, including ROCO, MedpixCap, IU-Xray, and MIMIC-CXR. **b,** Different types of medical text generation cases, including image caption and radiology report. **c,** Comparison of performance between RadFound and Med-Flamingo on ROCO and MedpixCap datasets for image caption measured by ROUGE-L or METEOR. **d,** Comparison of performance between RadFound and Med-Flamingo on IU-Xray and MIMIC-CXR datasets for radiology report generation measured by ROUGE-L or BLEU-4. **e,** An example case of MIMIC-Multi-Visit, including radiology images and reports from multiple previous visits. **f and g,** Comparison of performance between RadFound and Med-Flamingo on the MIMIC-CXR-visit under four settings: without previous reports or images, with previous reports only, with previous images only, and with both previous reports and images, measured by ROUGE-L (f) and BLEU-4 (g).

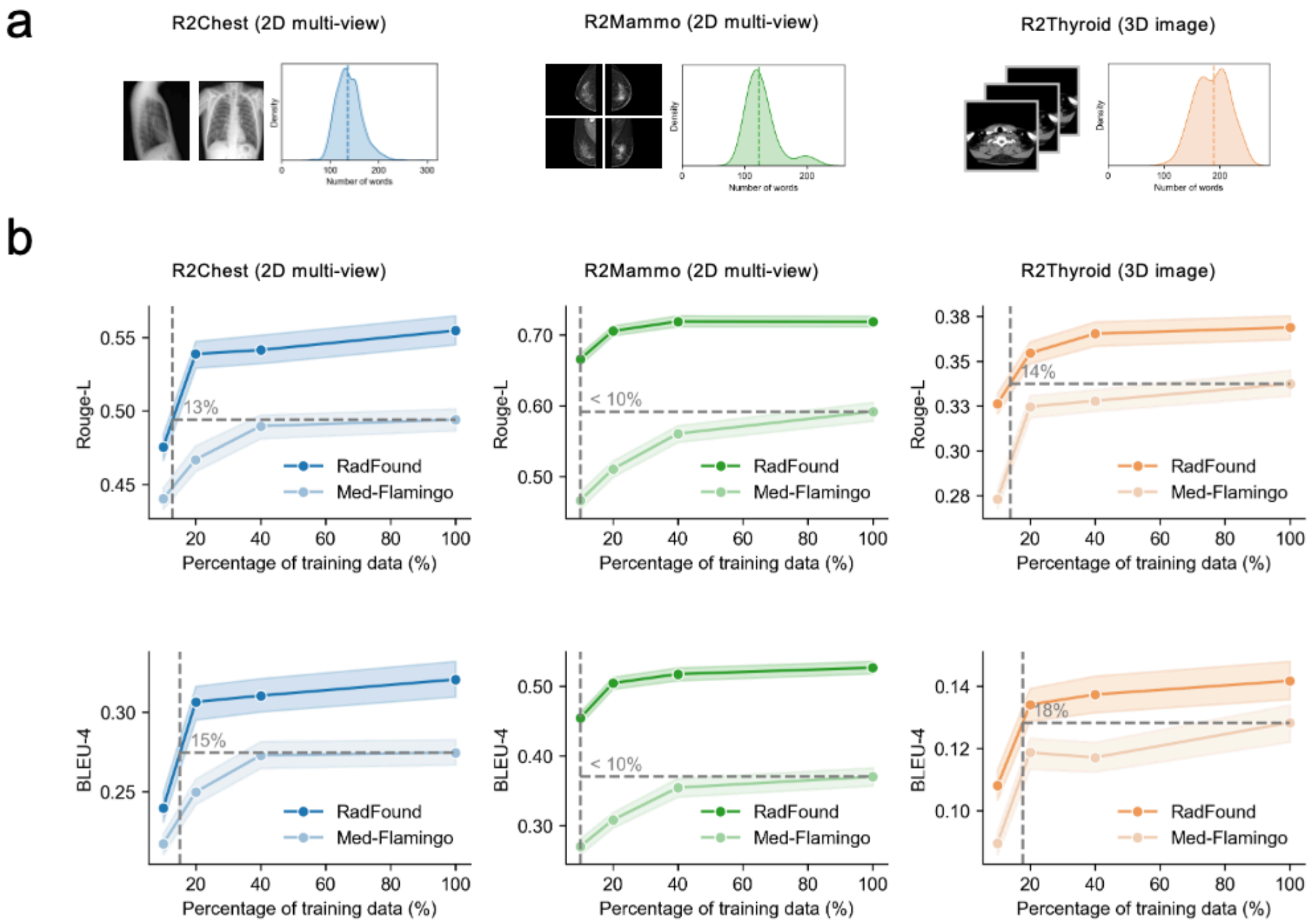

**Figure 4. Performance study of models for complex radiology report generation.** **a,** The distribution of textual length across three datasets and corresponding example cases, including R2Chest (2D multi-view), R2Mammo (2D multi-view), and R2Thyroid (3D image). **b,** The comparison of performance with varying training data sizes between RadFound and Med-Flamingo on three datasets: R2Chest (left), R2Mammo (middle), and R2Thyroid (right), measured by ROUGE-L (top) and BLEU-4 (bottom).

a

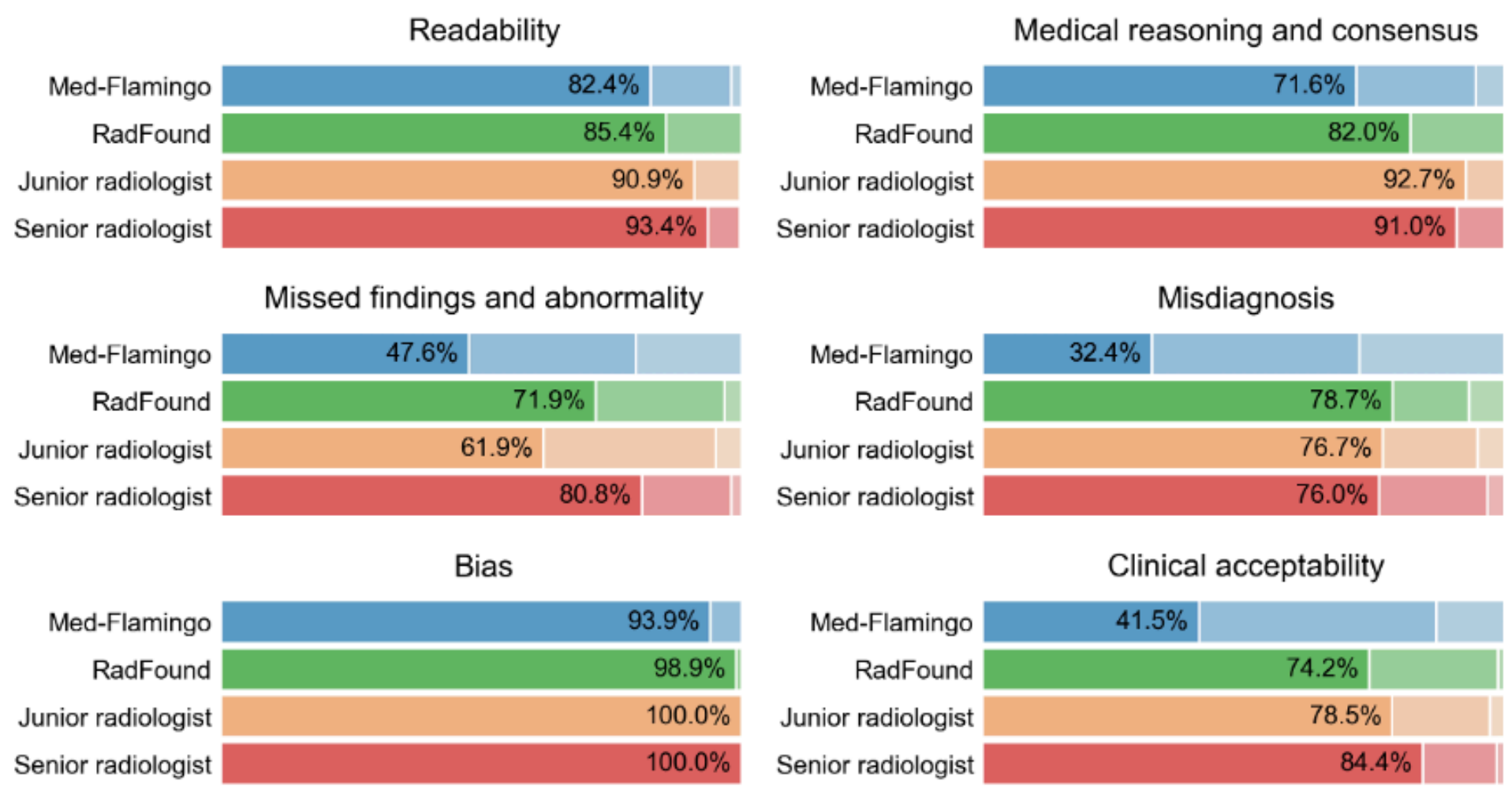

**Figure 5. Clinical evaluation between AI models and radiologists for real-world radiology report generation.**

**a,** The comparison between AI models and radiologists. The distribution of report evaluation scores averaged overall three datasets across six dimensions: Readability, Medical reasoning and consensus, Missed findings and abnormality, Misdiagnosis, Bias, and Clinical acceptability. Colors from dark to light represent from 5 points, 4-3 points, and 2-1 points, respectively.

## Supplementary Figure

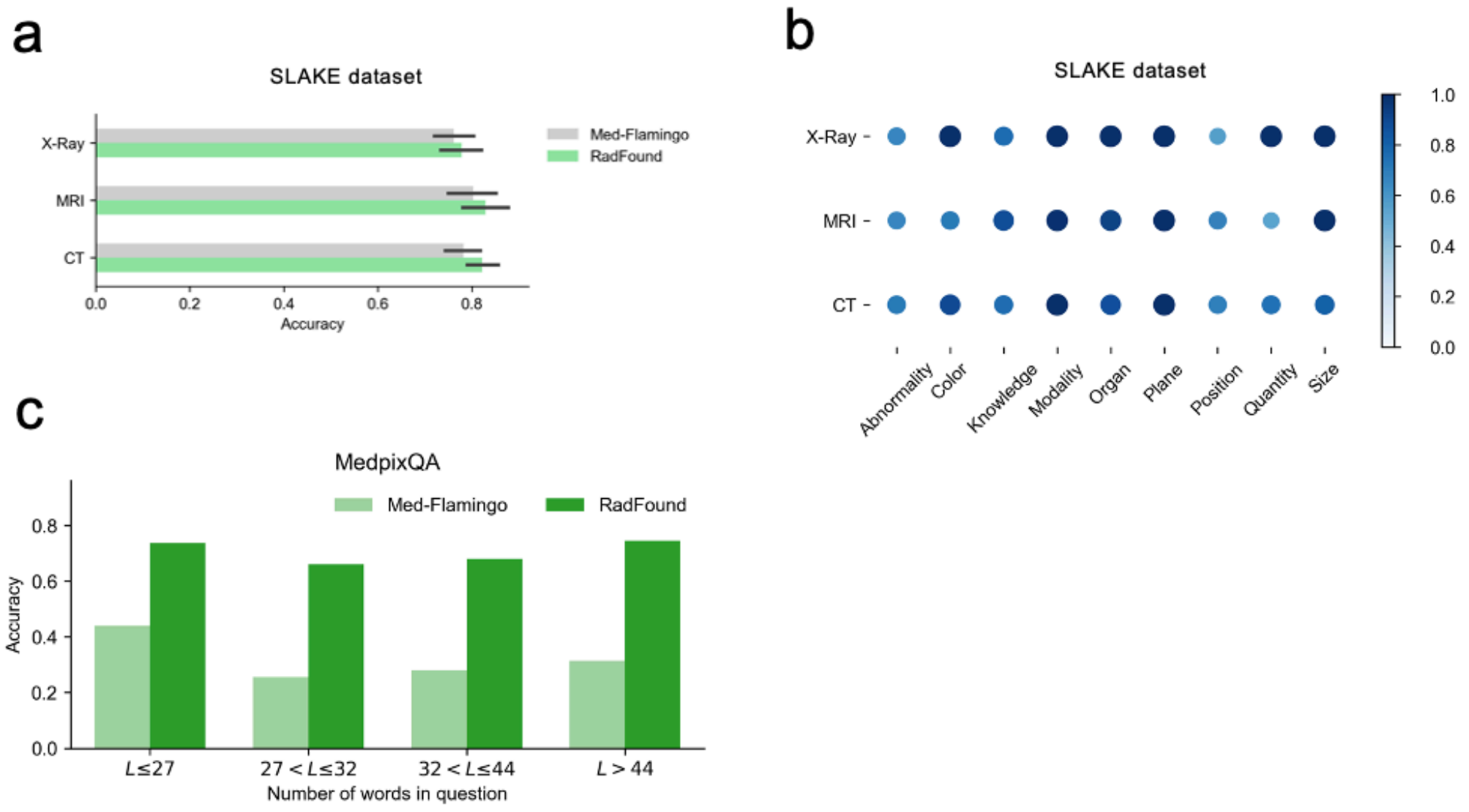

**Supplementary Figure 1. Performance of models for complex radiology report generation across organs, modalities, and question types.**

**a,** Comparison of performance between RadFound and Med-Flamingo on the SLAKE dataset across modalities. Green bar and grey bar represent the performance of RadFound and Med-Flamingo, respectively. **b,** Performance of RadFound on the SLAKE dataset across modalities and question types. The intensity of the color indicates the level of accuracy. **c,** Comparison of performance between RadFound and Med-Flamingo on the MedpixQA dataset across different question lengths. Dark green bar and light green bar represent the performance of RadFound and Med-Flamingo, respectively.

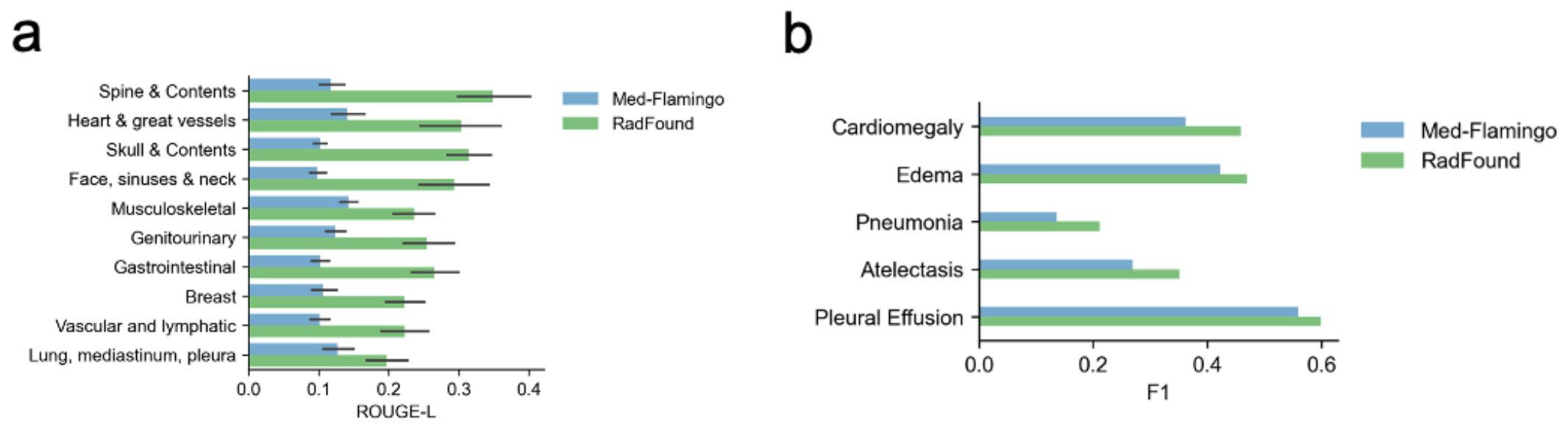

**Supplementary Figure 2. Performance of models for text generation across organs and clinical efficiency.**

**a,** Comparison of performance between RadFound and Med-Flamingo on the MedpixCap dataset across organs. Green bar and blue bar represent the performance of RadFound and Med-Flamingo, respectively. **b,** Clinical efficiency of RadFound on the MMIC-CXR dataset measured by F1-score of disease label detection. Green bar and blue bar represent the performance of RadFound and Med-Flamingo, respectively.